\documentclass[conference]{IEEEtran}
\IEEEoverridecommandlockouts    

\usepackage{cite}
\usepackage{amsmath,amssymb,amsfonts}
\usepackage{algorithm, algorithmic}
\usepackage{graphicx}
\usepackage{textcomp}
\usepackage{xcolor}
\usepackage{booktabs}
\usepackage{mathtools}
\usepackage{multirow}
\usepackage{url}
\usepackage[hidelinks]{hyperref}
\usepackage{orcidlink}
\usepackage{fontawesome}

\makeatletter
\let\MYcaption\@makecaption
\makeatother

\usepackage[font=footnotesize]{subcaption}

\makeatletter
\let\@makecaption\MYcaption
\makeatother

\def\BibTeX{{\rm B\kern-.05em{\sc i\kern-.025em b}\kern-.08em
    T\kern-.1667em\lower.7ex\hbox{E}\kern-.125emX}}

\DeclareMathOperator*{\argmin}{arg\,min}

\newcommand{\greenUp}{\textcolor{green}{\faArrowUp~}}
\newcommand{\redDown}{\textcolor{red}{\faArrowDown~}}
\newcommand{\redTimes}{\textcolor{red}{\faTimes~}}
\newcommand{\blackCheck}{\textcolor{black}{\faCheck~}}

\begin{document}

\title{\LARGE \bf A Collaborative Safety Shield for Safe and Efficient CAV Lane Changes in Congested On-Ramp Merging}

\author{
	\parbox{\textwidth}{%
		\centering
		Bharathkumar Hegde$^{1}$, M\'elanie Bouroche$^{1}$
	}%
	\thanks{$^{1}$School of Computer Science and Statistics, Trinity College Dublin, Ireland
		{\tt\small hegdeb@tcd.ie, Melanie.Bouroche@tcd.ie}}%
}

\maketitle
\thispagestyle{empty}
\pagestyle{empty}

\begin{abstract}
Lane changing in dense traffic is a significant challenge for Connected and Autonomous Vehicles (CAVs). Existing lane change controllers primarily either ensure safety or collaboratively improve traffic efficiency, but do not consider these conflicting objectives together.
To address this, we propose the Multi-Agent Safety Shield (MASS), designed using Control Barrier Functions (CBFs) to enable safe and collaborative lane changes. 
The MASS enables collaboration by capturing multi-agent interactions among CAVs through interaction topologies constructed as a graph using a simple algorithm.
Further, a state-of-the-art Multi-Agent Reinforcement Learning (MARL) lane change controller is extended by integrating MASS to ensure safety and defining a customised reward function to prioritise efficiency improvements. 
As a result, we propose a lane change controller, known as MARL-MASS, and evaluate it in a congested on-ramp merging simulation.
The results demonstrate that MASS enables collaborative lane changes with safety guarantees by strictly respecting the safety constraints. Moreover, the proposed custom reward function improves the stability of MARL policies trained with a safety shield. 
Overall, by encouraging the exploration of a collaborative lane change policy while respecting safety constraints, MARL-MASS effectively balances the trade-off between ensuring safety and improving traffic efficiency in congested traffic.
\end{abstract}



\section{Introduction}
Mitigating congestion is one of the major challenges in traffic flow management. 
The consequences of traffic congestion are widespread, and include less reliable travel times, additional fuel consumption, increased supply chain costs, inefficient traffic flow, and increased accident rates~\cite{shaygan_traffic_2022}. For example, the 2023 Urban Mobility Report estimates that the overall congestion cost in the United States is around \$224 billion~\cite{schrank_urban_2024}, resulting from a yearly value of 8.5 billion hours of traffic delay and 3.3 billion gallons of wasted fuel. 
In a congested merging scenario, efficient lane changes can improve the traffic flow to prevent or alleviate congestion. However, a lane change attempt in dense traffic is challenging, as it is a safety-critical task involving multiple CAVs making independent and concurrent decisions at high speeds. 

\begin{figure}[tbp]
\centering
    \begin{subfigure}{0.48\linewidth}
        \includegraphics[width=\linewidth]{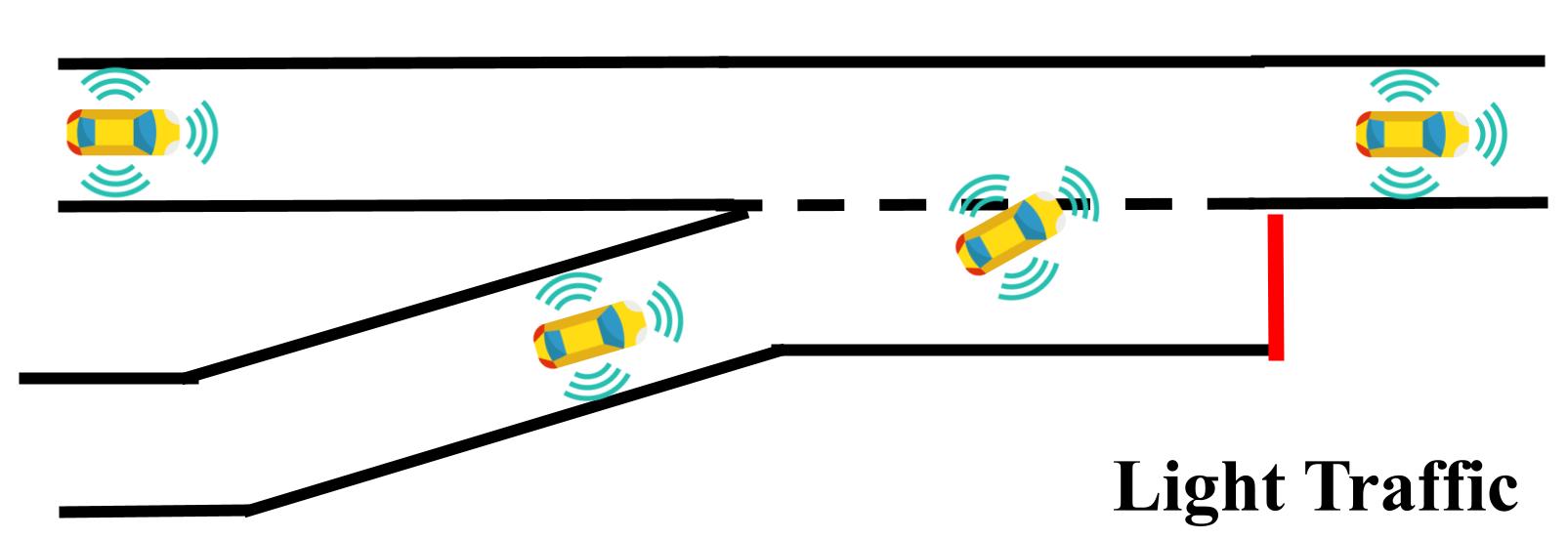}
    \label{fig:light_traffic}
    \vspace{-0.25cm}
    \end{subfigure}
    \begin{subfigure}{0.48\linewidth}
        \includegraphics[width=\linewidth]{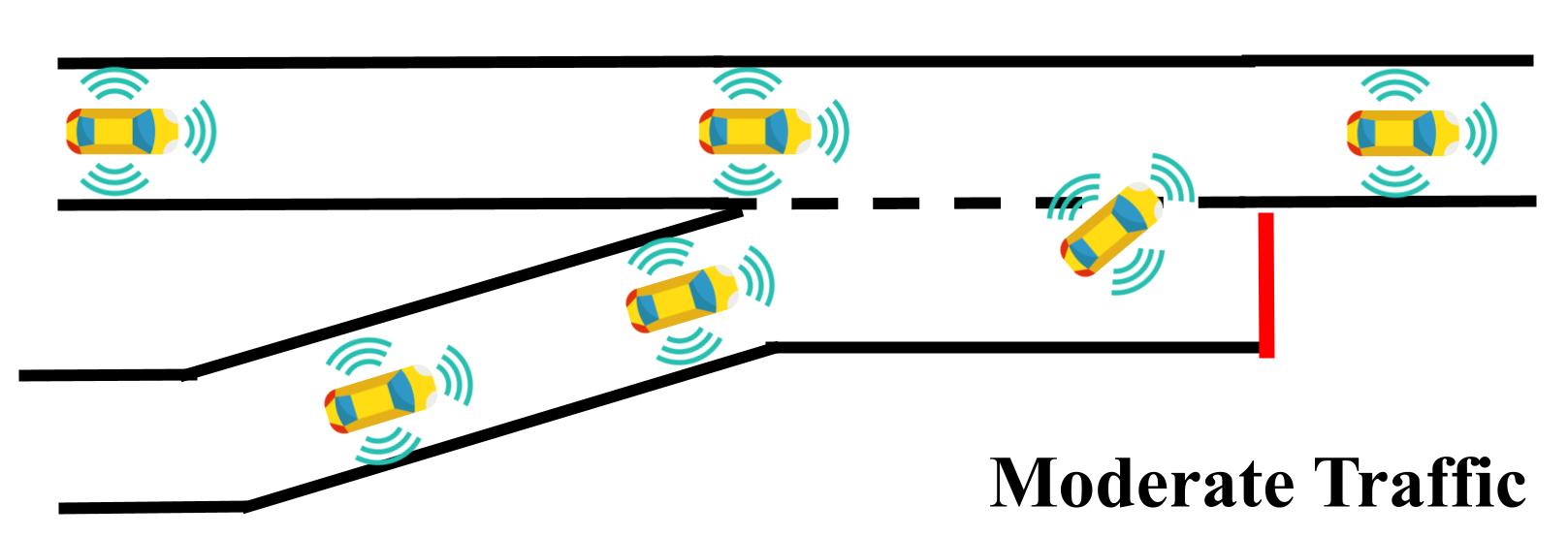}
    \label{fig:moderate_traffic}
    \vspace{-0.25cm}
    \end{subfigure}
    \\
    \begin{subfigure}{0.48\linewidth}
        \includegraphics[width=\linewidth]{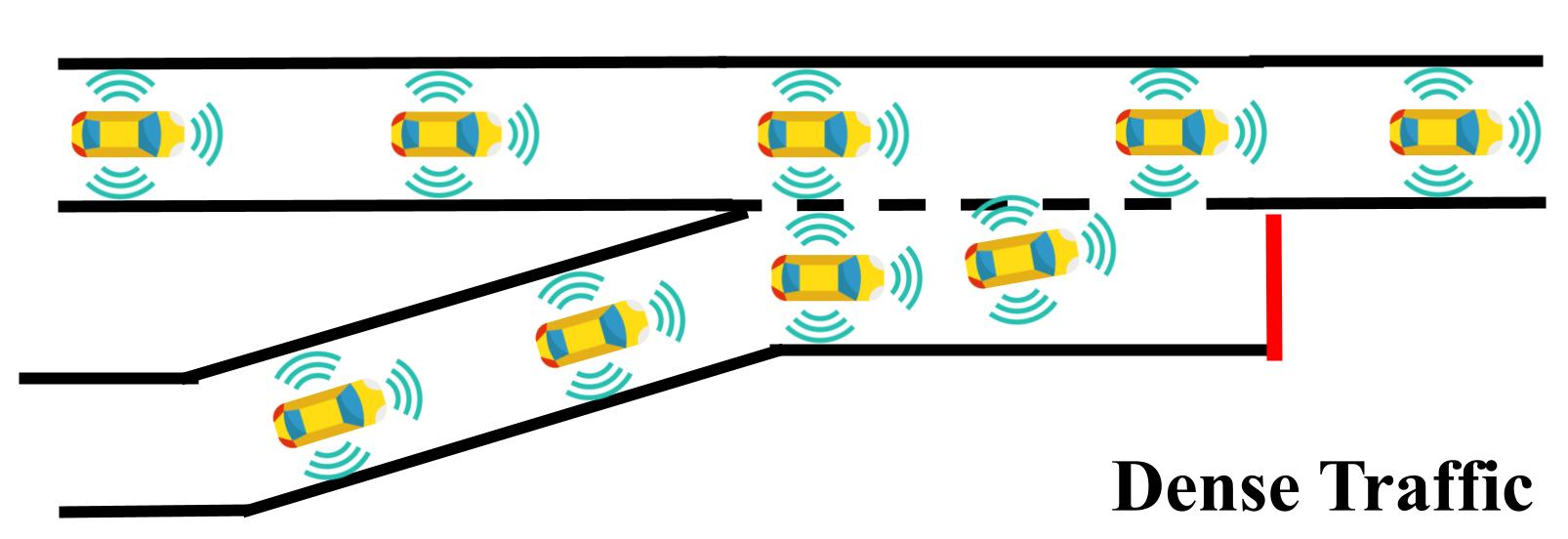}
    \label{fig:dense_traffic}
    \end{subfigure}
    \vspace{-0.75cm}
    \caption{Illustration of different traffic levels in a highway merging section}
    \label{fig:traffic_density_levels}
    \vspace{-0.65cm}
\end{figure}

Improper lane changes can result in a collision that could cause significant damage and loss of lives. Importantly, the safety of a vehicle depends not only on its control decisions, but also on the selfish decisions of surrounding vehicles that could cause a collision.
For example, it is hard to find an opportunity for merging in the dense traffic compared to light and moderate traffic, as illustrated in Fig.~\ref{fig:traffic_density_levels}.
In such scenarios, Connected Autonomous Vehicles (CAVs) can perform collaborative lane changes to improve traffic efficiency. As a result, CAV technologies have gained the attention of the research community as they enable vehicles to make collaborative control decisions using vehicular communication (V2X)~\cite{garg_can_2023}. 

To enable CAV collaboration, a multi-agent form of Reinforcement Learning (RL), known as Multi-Agent Reinforcement Learning (MARL), is gaining popularity in the research community due to its potential in learning a control policy for multiple agents acting concurrently and independently in a highly dynamic environment~\cite{hegde_design_2022, albrecht_multi-agent_2024}. Many forms of MARL have been applied for designing lane change controllers, including Deep Q-Networks (DQNs)~\cite{yu_distributed_2020},~\cite{dong_space-weighted_2021},~\cite{chen_graph_2021}, and Actor-Critic Networks (ACN)~\cite{zhou_multi-agent_2022},~\cite{hou_decentralized_2021},~\cite{chen_deep_2023}. 
Among them, the MARL lane change controller with a Centralised Shield, known as MARL-CS, uses the Multi-Agent extension of the Proximal Policy Optimisation (MAPPO) to train a policy using ACN. The MARL-CS is known for learning stable and reproducible policies to make efficient lane change decisions in a highway merging scenario~\cite{chen_deep_2023}. Moreover, this lane change controller is available in an open-source repository. Therefore, MARL-CS is considered a state-of-the-art MARL lane change controller for CAVs. 
While the primary objective of these MARL lane change controllers is to improve efficiency and minimise collisions, they do not provide safety guarantees.

Control Barrier Functions (CBFs) have been identified as a suitable option to ensure the safety of the decentralised MARL policies in CAVs~\cite{hegde_multi-agent_2024}. Indeed, as MARL policies can make unsafe control decisions, safety constraints defined using CBFs can be used to override these unsafe actions, ensuring that an agent remains in safe states~\cite{cheng_end--end_2019, zhang_control_2025}. 
The Hybrid Safety Shield (HSS) has been proposed to ensure the safety of MARL CAV lane change controllers, which combines an optimisation and a rule-based approach to effectively constrain the acceleration and steering angle inputs~\cite{hegde_safe_2025}. The HSS ensures the safety of a vehicle, despite independent and concurrent lane changes, by taking a cautious approach that assumes other vehicles make worst-case control decisions. 
The MARL lane change controller, integrated with HSS, namely MARL-HSS, improves traffic efficiency in light to moderate traffic conditions while ensuring safety. MARL-HSS, however, struggles to converge to a stable policy as traffic becomes heavier, due to limited opportunities for safe lane changes, resulting in poor traffic efficiency.

Instead, CAVs can improve traffic efficiency even in dense traffic through collaborative control decisions. 
Considering CAVs as Multi-Agent Systems (MAS), efficient lane change decisions can be made with CAV collaboration approaches, such as slot-based~\cite{marinescu_-ramp_2012}, game theory~\cite{jing_cooperative_2019}, and priority-based~\cite{chen_deep_2023, valiente_robustness_2022}. 
These methods, however, depend on a centralised server to orchestrate collaboration among vehicles in on-ramp merging, which limits their applicability and scalability. 
Alternatively, decentralised collaboration methods usually capture multi-agent dynamics through graphs, known as interaction topologies~\cite{butler_collaborative_2024, tan_distributed_2022, liang_mas-based_2023}. The potential application of this approach to design a safety shield for CAVs remains to be investigated.

In this paper, we explore the application of interaction topologies to develop a collaborative safety shield and integrate it with a MARL lane change controller to mitigate congestion in on-ramp merging while ensuring safety.
The main contributions are as follows:
\begin{itemize}
    \item We propose a novel collaborative safety shield, referred to as Multi-Agent Safety Shield (MASS), that builds on HSS to consider the control decisions of other CAVs to evaluate safe control inputs using CBF constraints.
    The MASS is integrated with MARL to define a collaborative controller for CAVs known as MARL-MASS, which is available at \url{https://github.com/hkbharath/MARL-MASS}
    
    \item To facilitate collaboration in MASS, we outline an algorithm to construct the interaction topology for CAVs in the highway merging scenario. This topology is represented as a graph that is used to evaluate joint safe control actions.
    
    \item A customised reward function is explicitly defined for the MARL controller integrated with MASS to prioritise traffic efficiency optimisation, while the safety shield ensures safety. 
    
    \item The MARL-MASS is empirically evaluated in a gym-like simulation platform to demonstrate its superiority over baseline MARL controllers in mitigating dense traffic while respecting the safety constraints.
\end{itemize}



\section{Background}
This section presents the background knowledge on vehicle control, MARL, and CBFs to establish a context for the proposed method.

\subsection{Vehicle control}
\label{subsec:vehicle_model}
Typically, a vehicle control has multiple hierarchical levels corresponding to different driving tasks~\cite{paden_survey_2016}. Among them, the behavioural layer makes discrete decisions to change lanes or follow the current lane. The motion planning layer then identifies the control references to execute the desired driving manoeuvre to implement each behavioural decision. These driving manoeuvres are then actuated by the vehicular controller based on a vehicle model that defines Autonomous Vehicle(AV) dynamics.

AV dynamics can be affected by a variety of factors ranging from vehicle control to weather conditions, which makes it challenging to model. For simplicity, the kinematic bicycle model is often used to model AV motion~\cite{polack_kinematic_2017}, and we demonstrate the proposed safety shield using this model. 
\begin{figure}[tbp]
    \centering
    \includegraphics[width=0.8\linewidth]{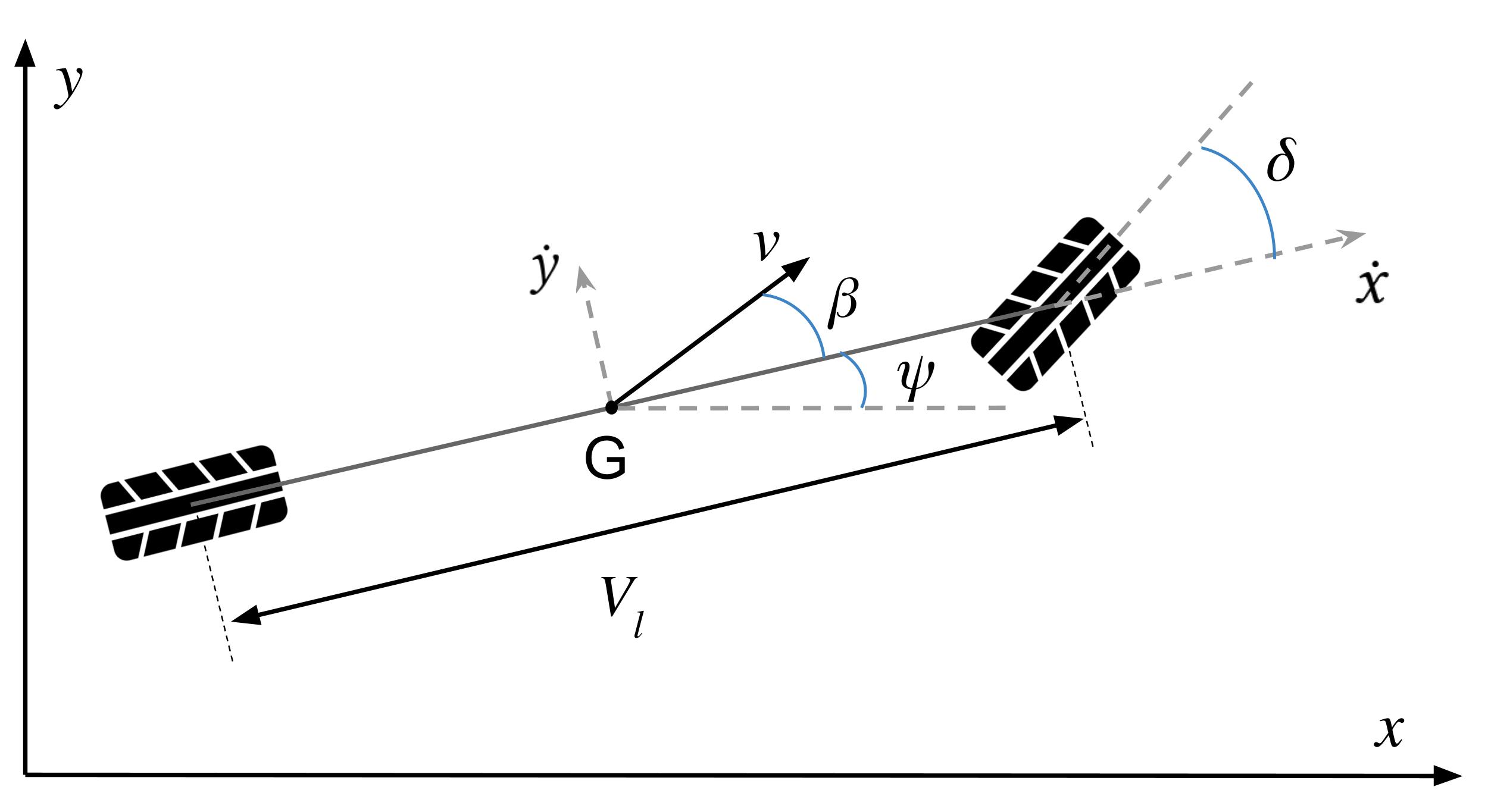}
    \caption{Kinematic bicycle model}
    \label{fig:kinematic_model}
    \vspace{-0.5cm}
\end{figure}
The kinematic bicycle model considers the two front wheels as one wheel and the same for the back wheels, as illustrated in Fig.~\ref{fig:kinematic_model}. The following equations define the vehicle model to update the state variables of a vehicle,
\begin{align}
    \label{eq:kinematic_model}
    \begin{split}    
    \dot{x} = v_x,~~~~
    \dot{y} = v_y,~~~~
    \dot{v}_x = a\cos{(\psi + \beta)},\\
    \dot{v}_y = a\sin{(\psi + \beta)},~~~~
    \dot{\psi} = \frac{2~v}{V_l}\sin{\beta},~~~~
    \end{split}
\end{align}
where $\beta$ is a slip angle at the centre of gravity evaluated based on the steering angle input $\delta$:
\begin{align}
    \label{eq:slip_angle}
    \beta &= \tan^{-1} \left( \frac{1}{2}\tan\delta \right)
\end{align}

\subsection{Multi-Agent Reinforcement Learning for lane changing}

In a state-of-the-art MARL lane change controller for on-ramp merging, the MARL-CS, lane change decisions are made as a discrete behavioural decision, such as \{\verb+right+, \verb+left+, \verb+follow lane+, \verb+speed up+, \verb+slow down+\}, based on the observed state space consisting of the following state variables,
$x$~: The longitudinal position of the vehicle,
$y$~: The lateral position of the vehicle,
$v_x$~: The longitudinal velocity of the vehicle, 
$v_y$~: The lateral velocity of the vehicle, and
$\psi$~: The vehicle heading with respect to the road.
At each discrete time step $t$, CAVs make a joint behavioural decision by combining the individual decisions.

The reward function used in MARL-CS rewards an agent for achieving individual and collaborative goals. 
The reward for a CAV $i$ is defined as
\begin{equation}
    r_{i} = w_\text{c} r_\text{c} + w_\text{s} r_\text{s} + w_\text{h} r_\text{h} + w_\text{m} r_\text{m}
    \label{eq:reward}
\end{equation}
where a CAV is rewarded for avoiding collision with $r_\text{c}$, maintaining desirable speed with $r_\text{s}$, maintaining desirable headway with $r_\text{h}$, and minimising the merging wait time with $r_\text{m}$. Each of these rewards is multiplied by a weight $w_*$ that can be tuned to prioritise the different objectives.
To reward collaborative goals, the individual rewards from the $n$ observed vehicles within the perception range are combined, resulting in the following overall reward function:
\begin{equation}
    R_{i} = \frac{1}{n}\sum_{j=0}^{n} r_{j}
\end{equation}

\subsection{Control Barrier Functions for motion planning}
CBFs are typically used to constrain a non-linear control system, such as a robotic controller or a CAV, to ensure safe control decisions~\cite{ames_control_2019},~\cite{han_multi-agent_2023}.
Consider a discrete time non-linear control system that defines the next state $s_{t+1}$ based on unactuated dynamics $f:\mathcal{S}\rightarrow \mathcal{S}$, and actuated dynamics $g: \mathcal{S} \rightarrow \mathbb{R}^{n,m}$, where $n$ and $m$ are the number of variables in the state space $\mathcal{S}$ and action space $U$, respectively. This transition dynamics can be defined as follows,
\begin{equation}
    \label{eq:nonlin_control}
    s_{t+1} = f(s_t) + g(s_t)u_t
\end{equation}
where $s_t \in \mathcal{S}$ is the system state, and $u_t \in U$ is the control action at time step $t$. 
The $f$ and $g$ are defined based on known system dynamics, and they are locally Lipschitz continuous, in other words, continuous functions limited by a maximum rate of change. 

Using continuously differentiable CBF $h: \mathcal{S} \rightarrow \mathbb{R}$, the safe set $C: \{s_t \in \mathcal{S}: h(s_t) \geq 0\}$ can be defined, where $h(s_t)$ is an affine barrier function of the form $h(s_t) = p^{\text{T}}s_t+q$ that defines a safety constraint for the non-linear control system.
Using this constraint, an optimisation problem can be formulated to evaluate the minimum control action $u_t$ that ensures safety of the control system in Eq.~(\ref{eq:nonlin_control})~\cite{cheng_end--end_2019}. The detailed derivation of this optimisation problem can be found in~\cite{cheng_end--end_2019} and~\cite{ames_control_2019}.

Application of CBFs to ensure the safety of MARL lane change controllers for CAVs is demonstrated in MARL-HSS~\cite{hegde_safe_2025}.
In this method, the HSS overrides potentially unsafe low-level actions, such as $v_\text{e}$, evaluated for executing a high-level behavioural decision made by a MARL policy for the ego vehicle. 
Although HSS considers the state of the observed vehicles, they are assumed to make a worst-case control decision $v^{\text{wc}}_\text{o}$ for evaluating the minimum correction value $v^{\text{cbf}}$ to comply with a CBF constraint $h(s_t)$. This ensures that the safety constraints are honoured irrespective of the control decisions of surrounding vehicles. Therefore, the safe action $v^{\text{safe}}$ is defined as 
\begin{equation}
    \label{eq:safe_v}
    v^{\text{safe}} = 
    \underbrace{
    \begin{bmatrix}
        v_{\text{e}} &
        v^{\text{wc}}_{\text{o}}
    \end{bmatrix}}_{v}
    + 
    \underbrace{
    \begin{bmatrix}
        v^{\text{cbf}}_{\text{e}} &
        0
    \end{bmatrix}}_{v^{\text{cbf}}}
\end{equation}

In HSS, longitudinal safety constraint is defined using a CBF $h_{\text{ol}}(s_t)$ based on time headway threshold $\tau$ with respect to the observed leading vehicle, defined as follows
\begin{align}
    \label{eq:h_lon}
    \begin{split}    
    h_{\text{ol}}(x_t) &= \Delta x_{\text{ol}} - x^{\text{safe}}\\
    \end{split}
\end{align}
where $x^{\text{safe}} = \tau * v_{\text{e}}$ and $\Delta x_{\text{ol}}$ is the longitudinal distance between the ego vehicle and the observed leading vehicle. 
Similarly, HSS uses lateral safety constraints with CBF constraints $h_{\text{oal}}(s_t)$ and $h_{\text{oar}}(s_t)$ with respect to the leading and rear vehicle in the adjacent lane. 
The HSS allows lane change only if the lateral constraints are satisfied to ensure safe lateral control decisions~\cite{hegde_safe_2025}.


\section{Proposed Approach}
\label{sec:proposed_approach}
\begin{figure}[tbp]
    \centering
    \includegraphics[width=0.9\linewidth]{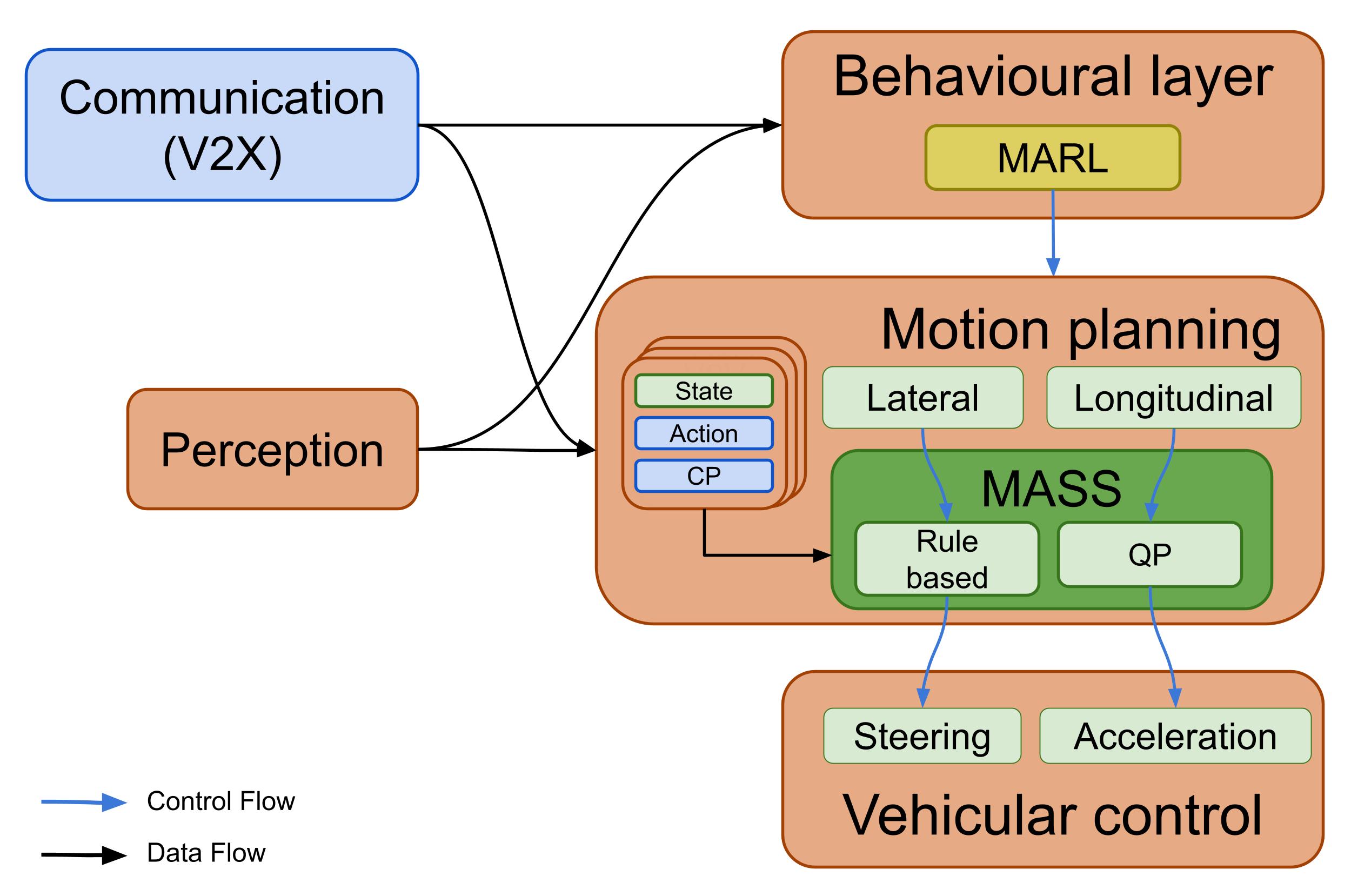}
    \caption{The MARL-MASS vehicle controller architecture}
    \label{fig:marl_mass}
    \vspace{-0.5cm}
\end{figure}

This section explains the main contributions of this paper, which result in MARL-MASS, a safe and efficient lane change controller in which the MARL behavioural layer is integrated with MASS, as illustrated in Fig.~\ref{fig:marl_mass}. The first contribution, MASS, is designed in Section~\ref{subsec:mass}. 
To identify interaction dependencies in MAS of CAVs, an algorithm is presented in Section~\ref{subsec:interaction_topology} to construct the interaction topology. 
As the collaborative safety shield ensures safety, MARL policy can focus entirely on efficiency improvements. Thus, Section~\ref{subsec:marl_reward} presents a suitable reward function for MARL lane change controllers with a safety shield.

\begin{figure*}[tbp]
    \centering
    \begin{subfigure}{0.20\linewidth}
	\begin{subfigure}{\linewidth}
		\includegraphics[width=\linewidth]{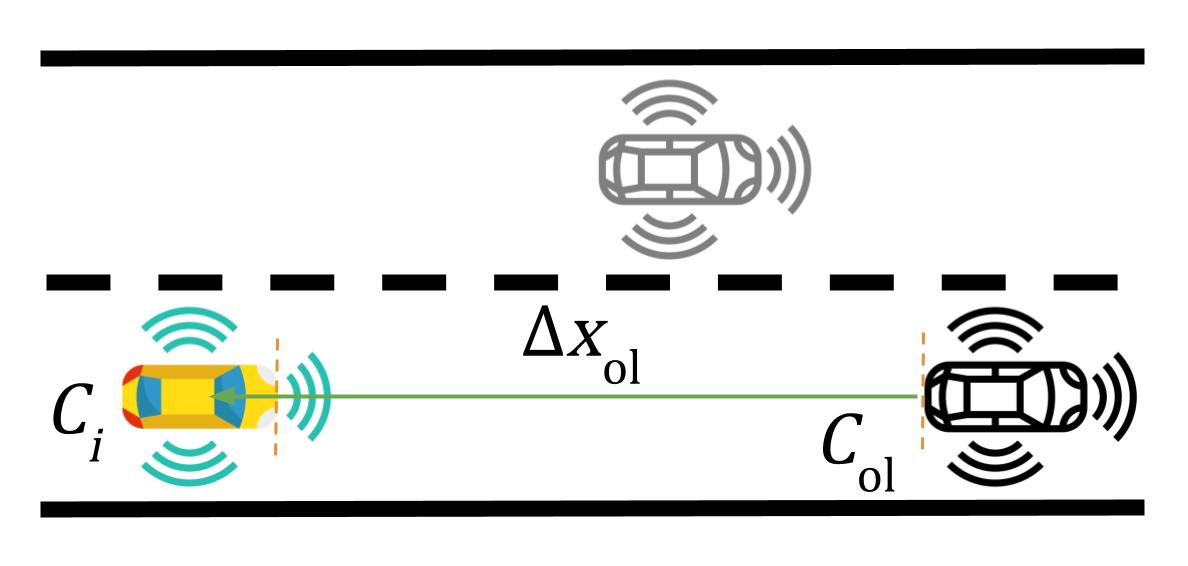}
		\caption{Follow lane}
		\label{fig:dep_graph_a}
	\end{subfigure}
    \end{subfigure}
    \begin{subfigure}{0.26\linewidth}
        \includegraphics[width=\linewidth]{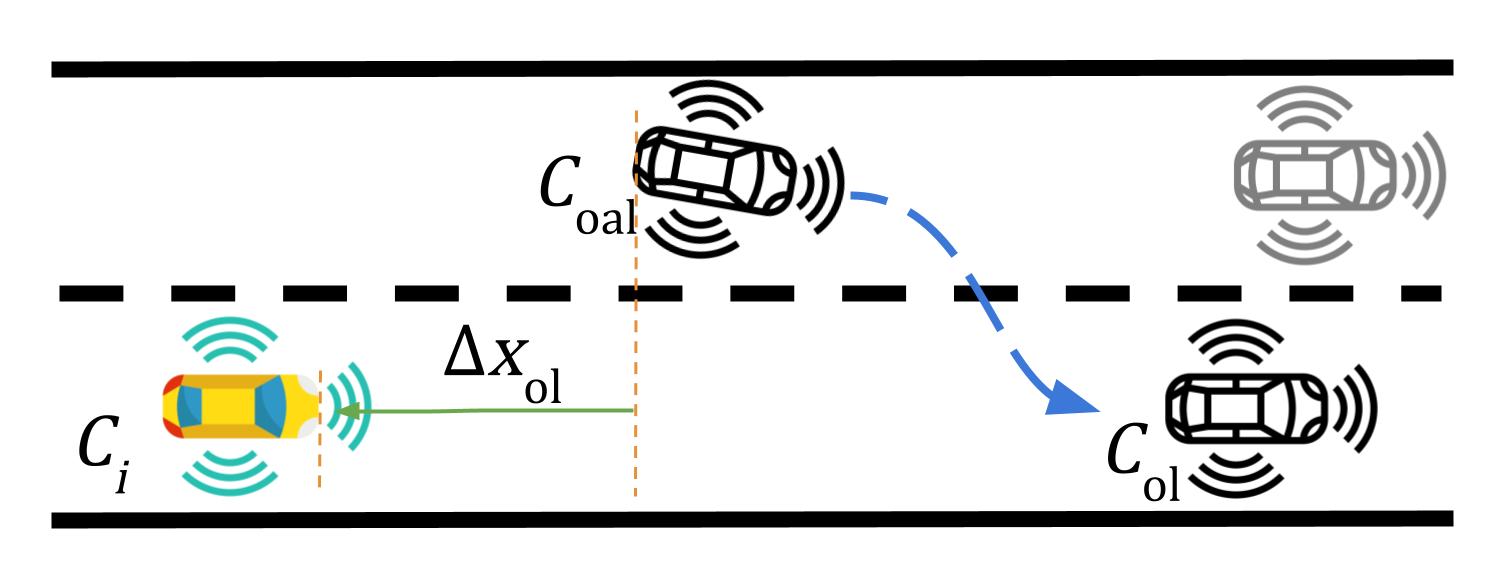}
        \caption{Adjacent vehicle changing lanes}
        \label{fig:dep_graph_b}
    \end{subfigure}
    \begin{subfigure}{0.28\linewidth}
        \includegraphics[width=\linewidth]{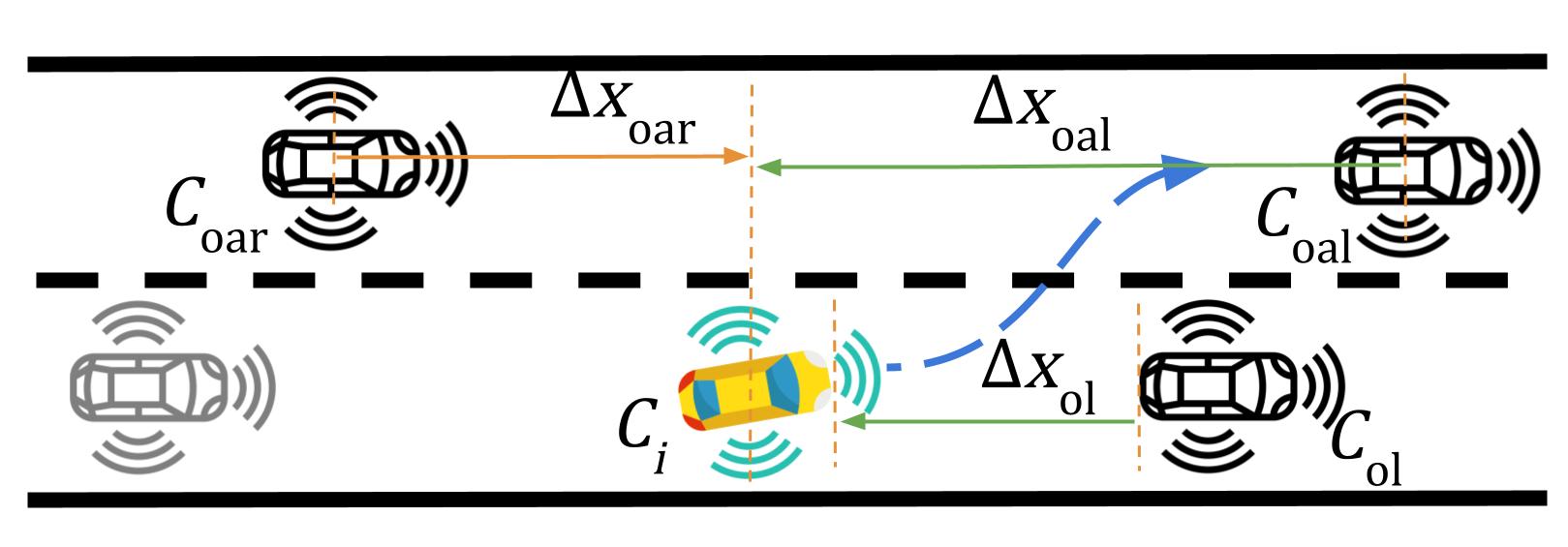}
        \caption{Lane changing vehicle}
        \label{fig:dep_graph_c}
    \end{subfigure}
    \begin{subfigure}{0.20\linewidth}
        \includegraphics[width=\linewidth]{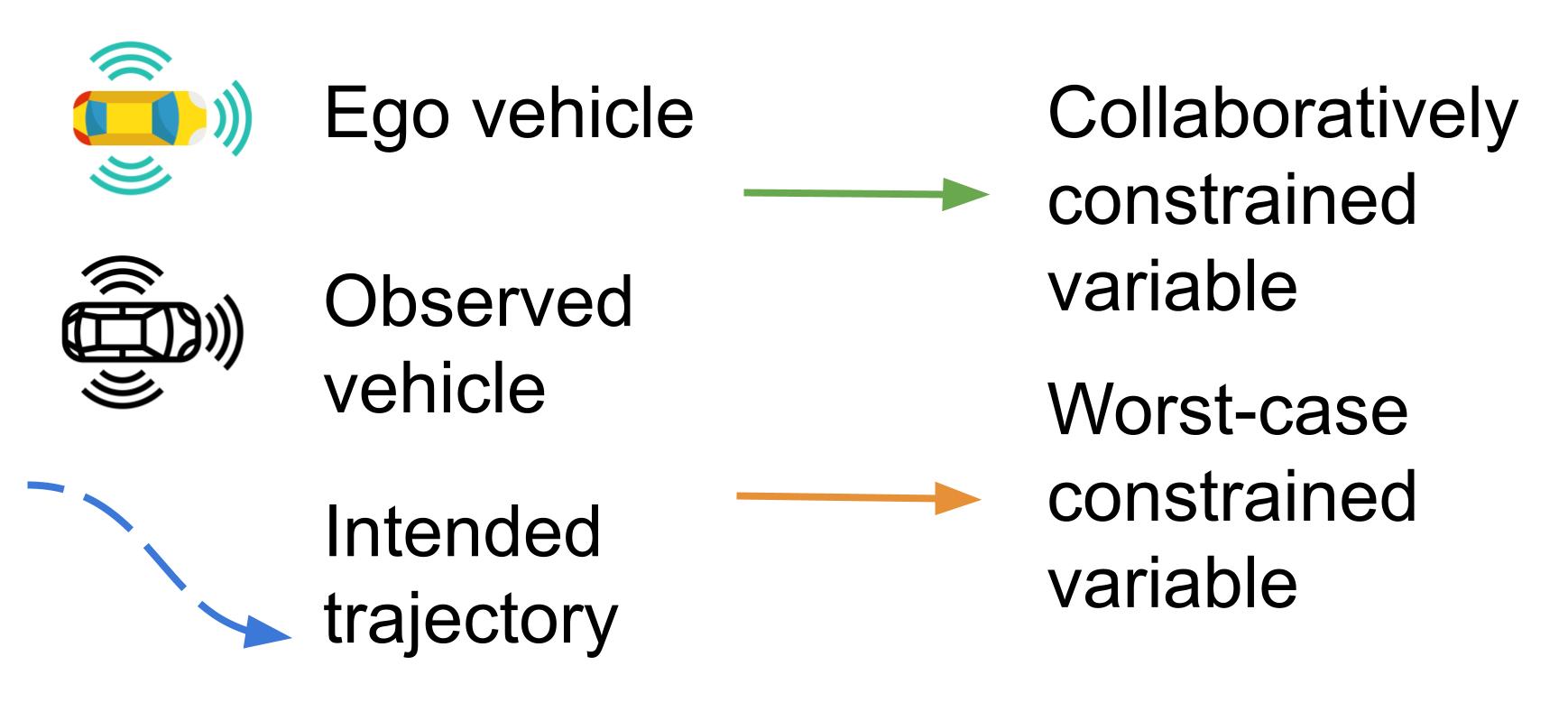}
    \vspace{0.2cm}
    \end{subfigure}
    \vspace{-0.25cm}
    \caption{Actor dependency for CAVs with decentralised CBF shield}
    \label{fig:dep_graph}
    \vspace{-0.5cm}
\end{figure*}

\subsection{Multi-Agent Safety Shield}
\label{subsec:mass}
MASS is a safety shield that enables CAVs to execute a behavioural action proposed by the MARL policy at the behavioural layer by making safe control actions at the motion planning layer at each time step. 
To formulate MASS, let us define a control system (\ref{eq:nonlin_control}) for the longitudinal motion of vehicles at a given time step as follows
\begin{align}
    \label{eq:nonlin_control_loncav_matrix}
    \begin{bmatrix}
    x_{\text{e}}\\
    x_\text{o}
    \end{bmatrix}
     =
    \underbrace{
    \begin{bmatrix}
         x_\text{e}\\
         x_\text{o}\\
    \end{bmatrix}}_{f(x)}
    +
    \underbrace{
    \begin{bmatrix}
         \cos{(\psi_\text{e} + \beta_\text{e})}\\
         g(x_o)\\
    \end{bmatrix}}_{g(x)}
    \underbrace{
    \begin{bmatrix}
        v_\text{e} & v^{\text{safe}}_\text{o}
    \end{bmatrix}}_{u_t}
\end{align}
where $\psi_\text{e}$ and $\beta_\text{e}$ are the ego vehicle's heading angle and slip angle defined in the vehicle model (\ref{eq:kinematic_model}). The $x_{\text{e}}$ and $v_{\text{e}}$ are the longitudinal position and velocity of the ego vehicle. Similarly, $x_{\text{o}}$ and $v^{\text{safe}}_{\text{o}}$ are the longitudinal position and velocity of an observed vehicle on which the ego vehicle depends for evaluating a safe control decision. The $g(x_{\text{o}})$ is the control parameter evaluated from the actuated dynamics of observed vehicles.
The ego vehicle depends on the observed vehicle to share the state $x_{\text{o}}$, safe control input $v^{\text{safe}}_{\text{o}}$, and the control parameter $g(x_{\text{o}})$. Although this control system is defined for a single observed vehicle, it can be further extended to consider multiple observed vehicles.

Since the state of the observed vehicle is dynamic, its safe control input $v^{\text{safe}}_{\text{o}}$ is considered in the control system.
The nominal control input $u_t$ can be considered as a multi-agent control input $v^{\text{ma}}$ as it consists of the nominal control input from the ego vehicle $v_{\text{e}}$ and the safe control input of the observed vehicle $v^{\text{safe}}_{\text{o}}$. The MASS in a CAV overrides the nominal control input $v_{\text{e}}$ in a minimally invasive manner to evaluate the safe control input  $v^{\text{safe}}$ as follows, 
\begin{equation}
    \label{eq:safe_v_ma}
    v^{\text{safe}} = 
    \underbrace{
    \begin{bmatrix}
        v_{\text{e}} &
        v^{\text{safe}}_{\text{o}}
    \end{bmatrix}}_{v^{\text{ma}}}
    + 
    \underbrace{
    \begin{bmatrix}
        v^{\text{cbf}}_{\text{e}} &
        0
    \end{bmatrix}}_{v^{\text{cbf}}}
\end{equation}

Using this control input, the following optimisation problem is formulated for a multi-agent CAV control system (Eq.~\ref{eq:nonlin_control_loncav_matrix}) to evaluate the minimum value for $v^{\text{cbf}}$ to ensure the safety of the ego vehicle.
\begin{align}
\label{eq:lon_cbf_opt_qp_ma}
\begin{split}
    v^{\text{cbf}} = & \argmin_{v^{\text{safe}}} ||v^{\text{safe}} - v^{\text{ma}}||_2 + K_{\epsilon}\epsilon~~~\text{s.t}~~Av^{\text{safe}} \leq b, \\
          \text{where } b =&~ [b_1, b_2,..., b_k], \\
          \text{~with}~ b_i =&~ p_i^{\text{T}}f(s_t)+ p_i^{\text{T}}(\eta - 1) s_t ~+\eta q_i + Av^{\text{ma}}
\end{split}
\end{align}

It is important to note that MASS depends on the state, the safe control input, and the control parameter from the observed vehicles to evaluate the safe control input.
This requirement creates a dependency between the observed vehicles and the ego vehicle. 
To handle such dependencies among CAVs, an interaction topology is constructed to facilitate the evaluation of a joint safe control input at the motion planning layer.

\subsection{Interaction Topology for CAVs}
\label{subsec:interaction_topology}

The interaction topology is defined for a MAS of CAVs crossing a merging section to identify the surrounding vehicles on which an ego vehicle depends to make its own safe control decisions. The surrounding vehicles are the subset of observed vehicles, such as the leading vehicle $C_{\text{ol}}$, the adjacent leading vehicle $C_{\text{oal}}$, and the adjacent rear vehicle $C_{\text{oar}}$.
The interaction topology is a graph $\mathcal{G}$ represented as a list of each CAV $C_i$ mapped to a set of parent vehicles $\mathcal{P}$ on which $C_i$ depends.
The interaction topology enables collaborative control decisions in a MAS, and it is defined based on the behavioural actions of CAVs. 

If the ego vehicle $C_i$  is following a lane, its control action depends on the immediate leading vehicle $C_{\text{ol}}$ within the communication range, as illustrated in Fig.~\ref{fig:dep_graph_a}. With this dependency, the ego vehicle can make an optimal control decision, based on the control decision made by the leading vehicle $C_{\text{ol}}$, to maintain a safe distance $\Delta x_{\text{ol}}$. 
If the adjacent vehicle $C_{\text{oal}}$ is attempting a lane change to join the current lane before the leading vehicle $C_{\text{ol}}$, the ego vehicle is considered to be dependent on the adjacent vehicle, as illustrated in Fig.~\ref{fig:dep_graph_b}. In this case, the ego vehicle considers the adjacent vehicle $C_{\text{oal}}$ to be the leading vehicle to maintain a safe longitudinal leading distance $\Delta x_{\text{ol}}$. This dependency allows collaboration among adjacent vehicles such that a vehicle is allowed to complete a lane change safely if the manoeuvre has already been initiated. Note that the safety shield allows initiating a lane change only when a safe gap exists in the target lane. Therefore, this dependency ensures that the decision from the vehicle attempting a lane change gets priority over the decision from the vehicle following a lane.

If the ego vehicle $C_i$ is changing lanes, then it is dependent on both the leading vehicle $C_{\text{ol}}$ and the adjacent leading vehicle $C_{\text{oal}}$ in the target lane as shown in Fig.~\ref{fig:dep_graph_c}. Based on the states and control actions from these vehicles, the ego vehicle can make safe control decisions to ensure that lane change manoeuvre is executed safely while maintaining safe longitudinal distances $\Delta x_{\text{ol}}$ and $\Delta x_{\text{oal}}$ with respect to the leading vehicle $C_{\text{ol}}$ and the adjacent vehicle $C_{\text{oal}}$ respectively. 
The rear vehicle in the adjacent lane $C_{\text{oar}}$ is not considered in the interaction topology to avoid deadlocks.  State variables of this vehicle, however, are used to constrain the ego vehicle from making unsafe lane changes based on lateral constraints and assuming the worst-case control decision.
This dependency enables a CAV to collaborate with surrounding vehicles to make control decisions that ensure safety while changing lanes. On reaching the target lane, the ego vehicle can continue to follow the lane while respecting the dependency defined earlier for a vehicle following a lane.

\begin{algorithm}[htbp]
\caption{Interaction topology}
\label{algo:interaction_topology}
\begin{algorithmic}[1]
\renewcommand{\algorithmicrequire}{\textbf{Input:}}
\renewcommand{\algorithmicensure}{\textbf{Output:}}
 \REQUIRE $\mathcal{N}$,~List of all vehicles
 \ENSURE  $\mathcal{G}$,~Interaction topology

 \STATE $\mathcal{G} \leftarrow \varnothing$
 \FORALL{$C_i$ in $\mathcal{N}$}
    
    \STATE $C_{\text{ol}} \leftarrow $ leading vehicle w.r.t $C_i$
    \STATE $C_{\text{oal}} \leftarrow $ adjacent leading vehicle w.r.t $C_i$

    \STATE $\mathcal{P} \leftarrow \{C_{\text{ol}}\}$
    
    \IF {$C_{\text{oal}}$ changing lane to current lane behind $C_{\text{ol}}$}
      \STATE $\mathcal{P} \leftarrow \{C_{\text{oal}}\}$
    \ENDIF
    
    \IF{$C_{i}$ is changing lane \AND $C_{\text{oal}}$ exists in the target lane}
        \STATE $\mathcal{P} \leftarrow \mathcal{P} \cup \{C_{\text{oal}}\}$
    \ENDIF
    
    \STATE $\mathcal{G} \leftarrow \mathcal{G} \cup \{ (C_i, \mathcal{P})\}$
 \ENDFOR
 \RETURN $\mathcal{G}$
\end{algorithmic}
\end{algorithm}

The steps to generate an integration topology are summarised in Algorithm~\ref{algo:interaction_topology}. This algorithm considers a list of $\mathcal{N}$ CAVs in the scenario as input and returns the interaction topology $\mathcal{G}$. This topology can be used by the MASS to ensure both longitudinal and lateral safety of CAVs individually by evaluating safety constraints relative to surrounding vehicles. By following the dependencies defined through interaction topology, the safety of all CAVs can be collectively ensured, because each CAV ensures safety with respect to its parent vehicles.

\subsection{Customised Reward for On-ramp Merging with MASS} 
\label{subsec:marl_reward}

The individual reward function is customised to ignore penalties for collisions and encourage CAVs to merge collaboratively, achieve higher speeds, and maintain smaller headway distances as the proposed safety shield (\ref{subsec:mass}) ensures that a safe headway is constantly maintained. To achieve this, we define the following customised reward function instead of the current reward function (\ref{eq:reward}) as follows
\begin{equation}
    r_{i} = w_\text{h} r_\text{h} + w_\text{s} r_\text{s} + w_\text{m} r_\text{m}
    \label{eq:new_reward}
\end{equation}

\begin{figure}[tbp]
    \centering
    \includegraphics[width=0.8\linewidth]{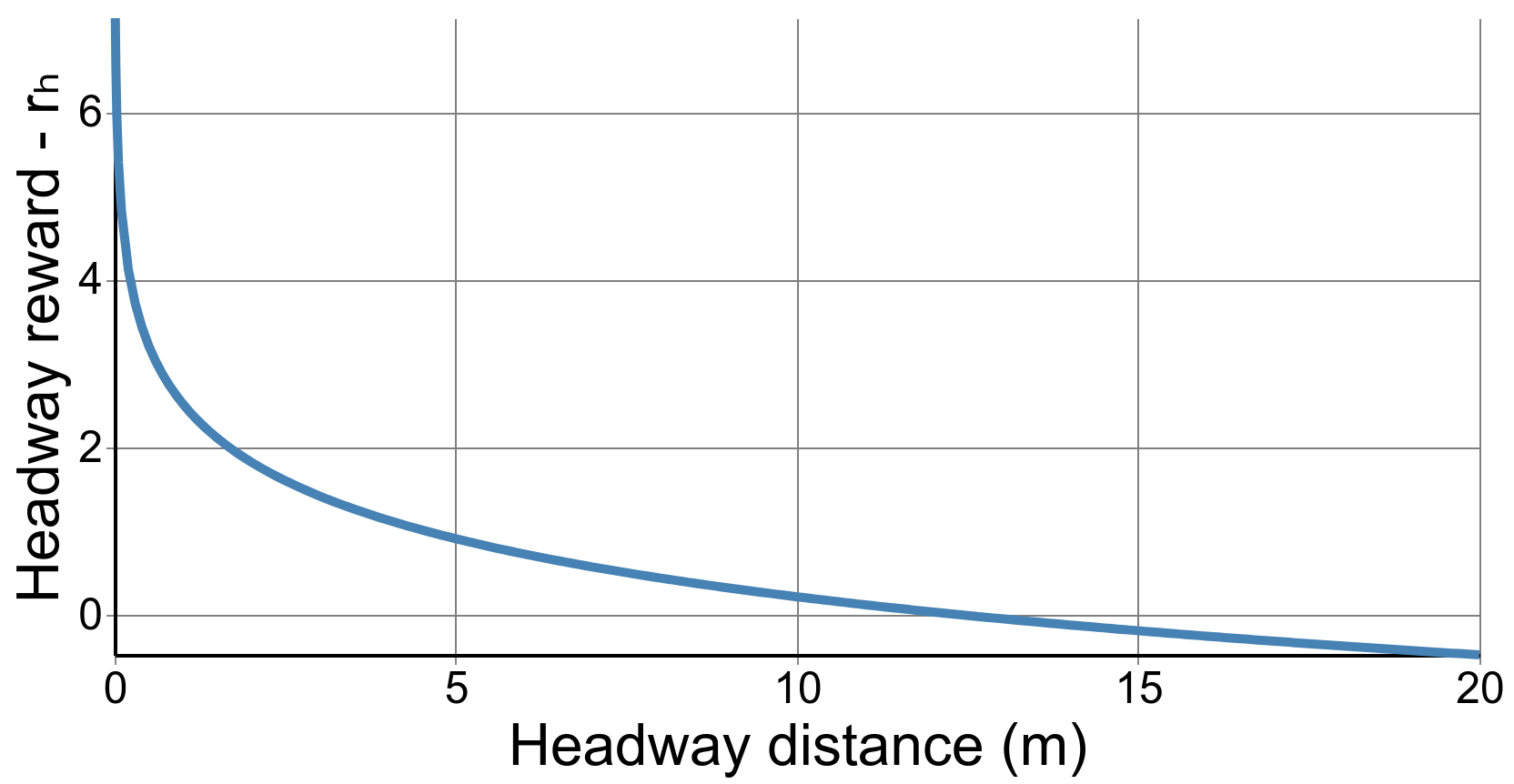}
    \caption{Illustration of the headway reward for a vehicle moving with a velocity of $25~\mathrm{m/s}$}
    \label{fig:headway_reward_plot}
    \vspace{-0.5cm}
\end{figure}

The headway reward $r_\text{h}$ is defined to penalise each agent for maintaining larger headway distances as follows
\begin{equation}
    r_\text{h} = - \log \frac{\Delta x_{\text{ol}}}{\tau * v_{\text{e}}}
    \label{eq:headway_reward}
\end{equation}
where $\Delta x_{\text{ol}}$ is the headway distance from the leading vehicle, $\tau$ is the desired time headway, and $v_{\text{e}}$ is the ego vehicle's velocity. 
This reward increases when the ego vehicle is closer to the leading vehicle and decreases when farther away. 
For instance, assuming a vehicle is moving at a velocity of $25~\mathrm{m/s}$ and considering a desirable time headway to be $\tau = 0.5~\mathrm{s}$, the headway reward $r_\text{h}$ is plotted as the headway distances change in Fig.~\ref{fig:headway_reward_plot}. This plot shows that agents can get higher rewards by maintaining a smaller headway.

The speed reward $r_\text{s}$ is evaluated as follows
\begin{equation}
    r_\text{s} = \min\left\{ \frac{v_{\text{e}} - v_{\text{min}}}{v_{\text{max}} - v_{\text{min}}}, 1\right\}
    \label{eq:speed_reward}
\end{equation}
where $v_{\text{min}}$ is the minimum and $v_{\text{max}}$ is the maximum desired speeds on highways. These values are chosen as $v_{\text{min}}= 10~\mathrm{m/s}$ and $v_{\text{max}}= 30~\mathrm{m/s}$  based on values used in MARL-CS~\cite{chen_deep_2023}. This reward encourages the agents to move at a higher speed close to $v_{\text{max}}$ to achieve higher traffic efficiency.

The merging reward $r_{\text{m}}$ penalises agents for waiting in the merging lane. This reward is defined as
\begin{equation}
    r_\text{m} = -\exp \left\{\frac{-(x_{\text{m}}-L)^2}{10L}\right\}
    \label{eq:merge_reward}
\end{equation}
where $x_{\text{m}}$ is the distance travelled by a vehicle on the merging lane and $L$ is the length of the merging lane. With this reward, the penalty increases as an agent travels farther in the merging lane without merging into the highway~\cite{chen_deep_2023}. 

Note that MASS restricts the vehicle from moving too close to the leading vehicle, but a vehicle may slow down to maintain a smaller headway distance to achieve a higher headway reward. However, slow speed is penalised through a speed reward $r_{\text{s}}$. Moreover, a vehicle would have to correctly pick the lane change opportunities to merge as soon as possible to avoid higher penalties from the merging reward.
Therefore, with this reward function, the agent is required to decide behavioural actions considering multiple objectives, such as higher speed, minimum headway distance, and efficient merging, based on traffic conditions.



\section{Evaluation and Results}

This section evaluates the proposed method, MARL-MASS, to demonstrate safety guarantees, analyse MARL training stability, and assess the trained policies.
Our lane change controller is compared with baseline MARL lane change controllers, such as \emph{Unsafe MARL}, \emph{MARL-CS}, and \emph{MARL-HSS}. The unsafe MARL lane change policy is trained without any shields to estimate the extent of the safety and efficiency benefits that can be achieved by the MARL. 
The MARL-CS and MARL-HSS lane change controllers integrate MARL with additional shields such as the centralised priority-based safety shield~\cite{chen_deep_2023}, and HSS~\cite{hegde_safe_2025}.
While MARL-HSS has demonstrated safety guarantees, MARL-CS improved traffic efficiency with collaborative lane changes.
Hence, these lane change controllers are used as baselines to evaluate the proposed collaborative safety shield, the MASS.

To evaluate these methods, first a simulation setup is defined for training MARL (Section~\ref{subsec:sim_setup}). Then, the safety guarantee from MASS is demonstrated through the minimum time headway observed during the MARL training (Section~\ref{subsec:safety_eval}). Next, the learning curves are illustrated to analyse the training stability of MARL policies with additional safety shields (Section~\ref{subsec:marl_stability}). Finally, policies trained with MARL-MASS, MARL-HSS, and MARL-CS are compared to analyse the trade-off between safety and efficiency.

\subsection{Evaluation scenario and simulation setup}
\label{subsec:sim_setup}

\begin{figure}[tbp]
\centering
    \includegraphics[width=\linewidth]{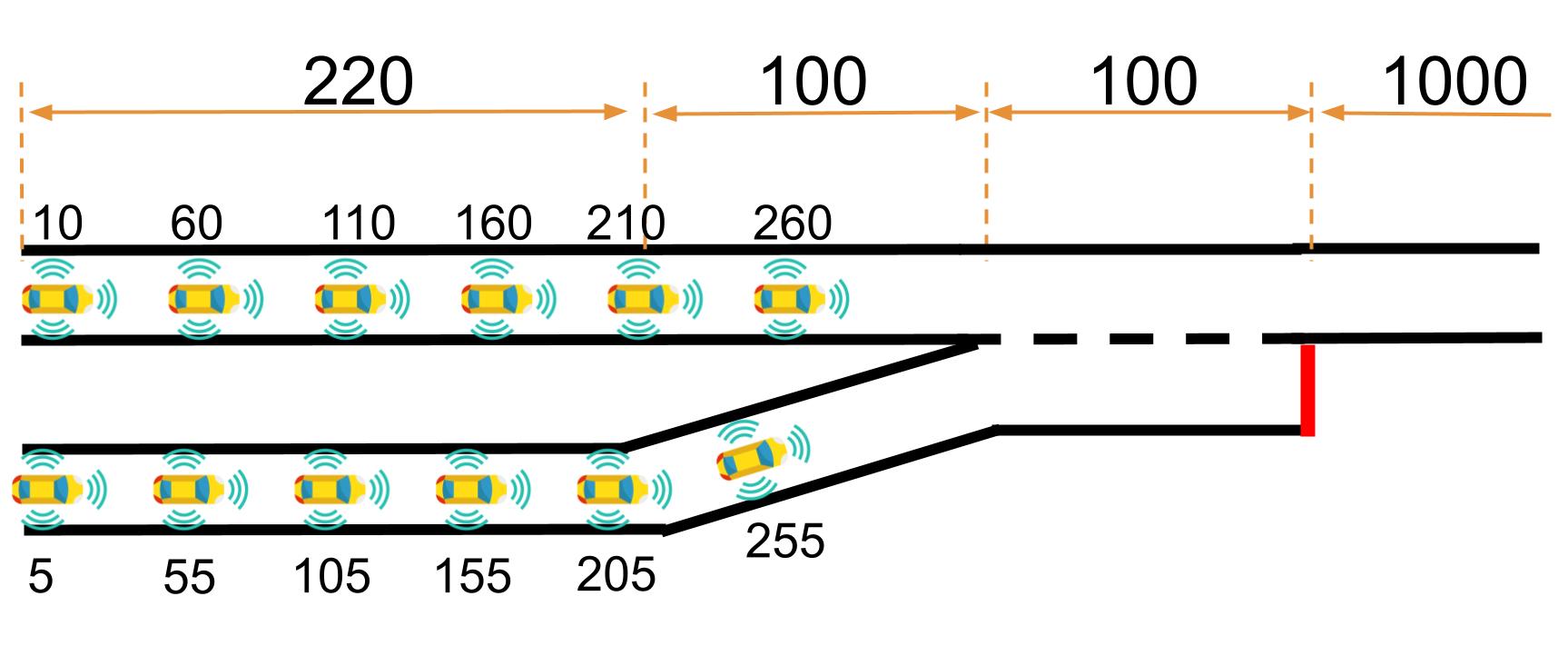}
    \vspace{-0.5cm}
    \caption{On-ramp merging scenario~\cite{hegde_safe_2025}}
    \label{fig:merging_scenario}
    \vspace{-0.5cm}
\end{figure}

An on-ramp merging scenario is used to train and evaluate the MARL lane change controllers. This is the same scenario used in the baseline MARL-HSS~\cite{hegde_safe_2025} with a higher traffic density consisting of 7-11 CAVs passing through a merging section of $100~\mathrm{m}$ as shown in Fig.~\ref{fig:merging_scenario}. The dense traffic is considered here, instead of light and moderate traffic considered in MARL-HSS, to showcase the MARL-MASS controller's ability to collaboratively create gaps that enable safe lane changes in congested traffic. In the MASS, CBF constraints are configured to maintain a $0.5~\mathrm{s}$ time headway with a leading vehicle, as CAVs are expected to maintain a smaller headway compared to AVs or human-driven vehicles. CAVs are assumed to communicate within a range of $180~\mathrm{m}$.

The \emph{highway-env} simulator~\cite{leurent_environment_2018}, which is widely used to evaluate AI-based AV controllers, is extended by integrating it with the MARL algorithm (MAPPO) and MASS to enable training MARL-MASS policies.
For training, the reward weights are set differently for the customised reward function as $w_\text{h}=1$, $w_\text{s}=4$, and $w_\text{m}=8$ to align with its priorities to improve efficiency. Specifically, $w_\text{m}$ is doubled to prioritise collaborative merging, $w_\text{s}$ is increased to reward higher speeds, and $w_\text{h}$ is reduced as maintaining a smaller headway is the third priority.
Although these weights can be configured for different driving modes, for simplicity, the same weights are used for all the vehicles in this analysis.
The MARL-MASS policies are trained using the learning rate $\alpha = 1e^{-4}$ with five random seeds to capture the variability of rewards. During the training for $20,000$ episodes, intermediate policies are evaluated using $20$ episodes for every $200$ training episodes. 
For training the lane change controllers considered in the evaluations, a shared Ubuntu 24.04 server with AMD EPYC 9654 processor, $4\times40$ GB GPUs (Nvidia-L40), and $1.5$ TB memory is used. The training, however, only utilised approximately $1.5$ GB GPU memory and $3$ GB RAM memory.
\vspace{-1pt}
\subsection{Safety evaluation}
\label{subsec:safety_eval}

\begin{figure}[tbp]
\centering
    \includegraphics[width=0.9\linewidth]{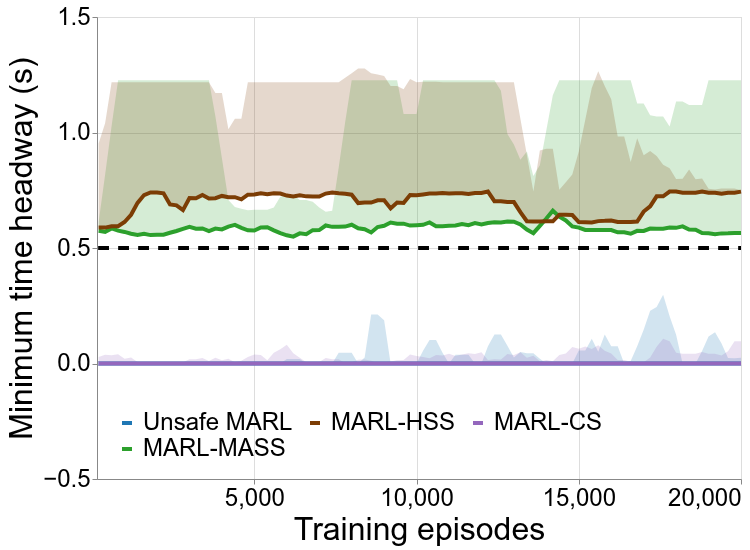}
    \vspace{-0.25cm}
    \caption{Minimum time headway observed during the MARL training in dense traffic. The spread illustrates the range (maximum and minimum) of minimum time headway observed over 5 random seeds.}
    \label{fig:min_hewadway}
    \vspace{-0.5cm}
\end{figure}

The safety guarantee from MASS is analysed in dense traffic at an on-ramp merging scenario that offers highly challenging situations with very few opportunities for safe lane changes. In Fig.~\ref{fig:min_hewadway}, the minimum time headways observed while training unsafe MARL, MARL-CS, MARL-HSS, and MARL-MASS are plotted. This parameter is captured by finding the minimum time headway from all CAVs at each step of the episodes used for the intermediate evaluations while training. 
As unsafe MARL and MARL-CS do not provide safety guarantees, they make unsafe control decisions that consistently violate the safety constraint, leading to crashes. 
These violations can be avoided with safety shields such as HSS and MASS, which provide safety guarantees by strictly respecting the safety constraint of a $0.5~\mathrm{s}$ time headway. 

While training MARL-HSS, a higher headway distance is observed compared to MARL-MASS, indicating inefficient road utilisation and a higher chance of congestion. To avoid congestion in this scenario, MASS enables CAVs to collaboratively create safe lane change opportunities. Moreover, the custom reward function further pushes the MARL policies to capitalise on these opportunities by maintaining a smaller time headway without violating the safety constraint.
The proposed safety shield, therefore, ensures that the safety constraint defined using the time headway is strictly enforced, even in challenging scenarios presented in a congested merging scenario.

\subsection{MARL training stability}
\label{subsec:marl_stability}

\begin{figure*}[tbp]
\centering
    \begin{subfigure}{0.9\linewidth}
    \begin{subfigure}{0.48\linewidth}
        \includegraphics[width=\linewidth]{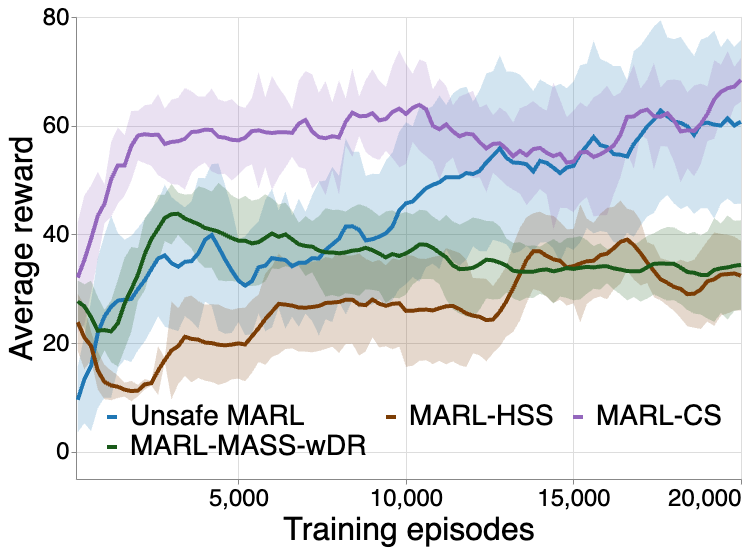}
    \caption{Reward curves from baseline controllers with the default reward}
    \label{fig:reward_curve_baselines}
    \end{subfigure}
    \begin{subfigure}{0.48\linewidth}
        \includegraphics[width=\linewidth]{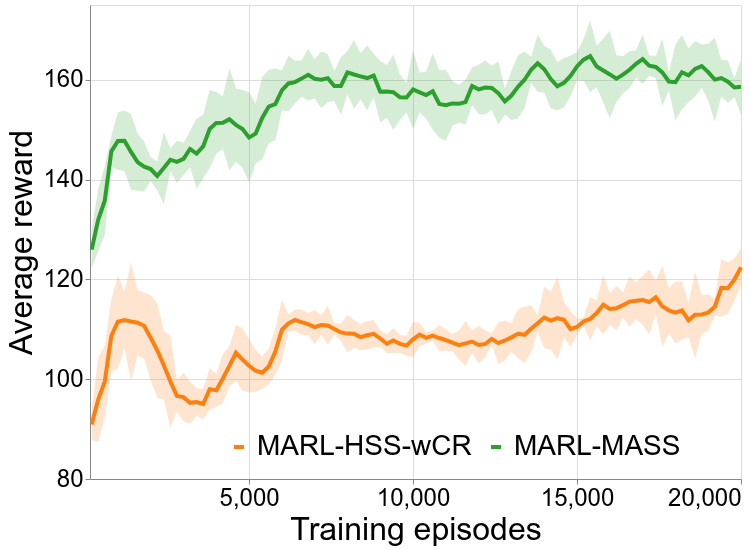}
    \caption{Reward curves with a customised reward}
    \label{fig:reward_curve_custom_reward}
    \end{subfigure}
    \end{subfigure}
    \caption{Comparison of the reward curves of training MARL lane change controllers. The spread indicates the standard error observed over 5 random seeds. The curves are smoothed using an average over a window of 5 evaluation epochs.}
    \label{fig:reward_curves}
    \vspace{-0.5cm}
\end{figure*}

The training stability of MARL is analysed by comparing the reward curves from the MARL lane change controllers with the default reward and the proposed reward function in Fig.~\ref{fig:reward_curves}, which is split into two plots to clearly demonstrate the superiority of the custom reward function and MASS.
The learning curves from policies trained using a default reward function (\ref{eq:reward}) are plotted in Fig.~\ref{fig:reward_curve_baselines}.
Although the unsafe MARL policy achieves rewards comparable to the MARL-CS policy at later training episodes, it exhibits higher variance, indicating instability without the centralised shield. 
As expected, the MARL-HSS policy fails to converge as it is challenging to find safe opportunities to change lanes in dense traffic.
Although MARL-MASS with default reward, referred to as MARL-MASS-wDR, achieves higher rewards at the initial stages of training, it converges towards a policy with a smaller reward. Moreover, a higher variance in the reward curve highlights the default reward function's inability to train a stable MARL-MASS policy that consistently achieves higher rewards. 

The reward curves from policy training that uses the proposed custom reward function~(\ref{eq:new_reward}) for lane change controllers integrated with a CBF safety shield, such as HSS or our MASS, are illustrated in Fig.~\ref{fig:reward_curve_custom_reward}. The MARL-HSS policy trained with a custom reward is referred to as MARL-HSS-wCR in this plot.
Using the proposed custom reward function, these MARL policies achieve a higher reward with less variance compared to the default reward.
While the higher reward is primarily due to a higher weight $w_s$ for the speed reward $r_s$, less variance indicates the suitability of the proposed reward function for training lane change controllers with a safety shield.

Compared to MARL-HSS, the proposed controller achieves significantly higher rewards with the proposed reward function, as shown in Fig.~\ref{fig:reward_curve_custom_reward}. 
Moreover, a consistently increasing learning curve with minimal variance during MARL-MASS policy training indicates stable convergence.
These observations highlight the increased safe lane change opportunities unlocked by the collaborative nature of MASS.

\subsection{MARL policy evaluation}
\label{subsec:policy_evaluation}

\begingroup
\renewcommand{\arraystretch}{1.5}
\begin{table}[tbp]
\caption{Policy evaluation using 100 episodes of congested merging scenarios. The number in the brackets is the standard error observed over policies trained with 5 different random seeds.}
\vspace{-0.5cm}
\begin{center}
\resizebox{\linewidth}{!}{%
\begin{tabular}{rlll}
\toprule
\textbf{} & \begin{tabular}[c]{@{}c@{}} Minimum \\ headway ($\mathrm{s}$)\end{tabular} & \begin{tabular}[c]{@{}c@{}} Speed ($\mathrm{m/s}$) \end{tabular} & \begin{tabular}[c]{@{}c@{}} Merging \\ percentage (\%)\end{tabular}\\ \midrule

\textbf{MARL-MASS (Ours)} & \blackCheck \textbf{0.59(0.01)} & \blackCheck \textbf{24.71(0.09)} & \blackCheck \textbf{83.36(0.38)} \\
\vspace{0.1cm}
{MARL-MASS-wDR} & \blackCheck 0.59(0.03) & \redDown 23.23(0.40) & \redDown 69.26(3.65) \\
 
{MARL-HSS-wCR} & \redDown 0.64(0.02) & \redDown 23.87(0.09) & \redDown 69.95(1.08) \\
 
{MARL-HSS} & \redDown 0.72(0.03) & \redDown 21.95(0.31) & \redDown 52.75(7.07) \\

{MARL-CS} & \redTimes 0.00 & \greenUp 25.83(0.84) & \greenUp 89.08(2.42) \\

{Unsafe MARL} & \redTimes 0.00 & \greenUp 26.26(1.52) & \redDown 76.91(14.96) \\
  
\bottomrule
\end{tabular}%
}
\vspace{-0.5cm}
\label{tab:policy_evaluation}
\end{center}
\end{table}

\endgroup



 
 

 
 

The best performing policies from each lane change controller, with the highest average reward, are evaluated using 100 \emph{policy evaluation episodes}, which are different from the episodes used for intermediate evaluation while training. 
These policies are compared to analyse traffic efficiency using average speed~($\mathrm{m/s}$), minimum headway~($\mathrm{s}$), and merging percentage. A higher average speed indicates higher traffic efficiency, whereas a lower minimum headway, close to $0.5~\mathrm{s}$, indicates efficient road space utilisation while respecting the safety constraint.  
The merging percentage is the percentage of vehicles spawned on the merging road that were successfully merged into the highway traffic. Since dense traffic offers only a few opportunities for safe lane changes, a higher merging percentage indicates a higher likelihood of collaborative lane changes, allowing more vehicles to merge into the highway.
The average values of these parameters over the policy evaluation episodes are presented in~Table~\ref{tab:policy_evaluation}.

Among the lane change controllers that ensure safety, the lowest time headway greater than $0.5~\mathrm{s}$ is observed with our lane change controller, indicating efficient road utilisation while ensuring strict compliance with safety constraints.
The advantages of training MARL using MASS and a custom reward (MARL-MASS), compared to MARL-HSS trained with a default reward, can be clearly observed in the gradual improvements in evaluation parameters of the intermediate policies, MARL-HSS with the custom reward (MARL-HSS-wCR) and MARL-MASS with the default reward (MARL-MASS-wDR), as shown in~Table~\ref{tab:policy_evaluation}.  
Although a small increase in the merging percentage from $52.75\%$ to $69.95\%$ is observed when using the customised reward function to train MARL-HSS policies, a significant increase from $69.95\%$ to $83.36\%$ is observed with MARL-MASS. The increase in merging percentage can be attributed to the collaborative nature of MASS, which slows the ego vehicle and creates safe gaps for merging.
Moreover, $12.57\%$ increase in average speed supplements the advantages of collaborative lane changing with MARL-MASS over MARL-HSS.

Conversely, lane change controllers without safety guarantees, such as unsafe MARL and MARL-CS, are observed to achieve slightly higher traffic efficiency than our approach, with higher average speeds. While the MARL-MASS policies exhibit a $6.54\%$ higher merging percentage compared to unsafe MARL, they lag behind MARL-CS by $5.72\%$.
These minor improvements, however, are realised at the cost of safety, as indicated by a minimum headway of $0.00~\mathrm{s}$, representing a collision.
Therefore, the proposed lane change controller is considered to achieve traffic efficiency comparable to unsafe baselines.

In sum, MARL-MASS policies are capable of mitigating congestion in on-ramp merging with dense traffic, as they improve traffic efficiency compared to policies trained with the baseline safety shield, HSS. 
Most importantly, MARL-MASS ensures safe lane change manoeuvres with traffic efficiency comparable to that of unsafe MARL and MARL-CS policies.


\section{Conclusion}
Performing efficient lane changes while ensuring safety in a congested merging section is a challenging task for CAVs. 
This paper proposes a lane change controller, known as MARL-MASS, that improves traffic efficiency using MARL and provides safety guarantees using a novel collaborative safety shield, denoted as MASS. To drive the collaboration among CAVs, we propose an algorithm to construct the interaction topology that captures dependencies among CAVs. Furthermore, for agents with a safety shield, we define a custom reward function that focuses on efficiency.   
The proposed method is evaluated by simulating a highway merging scenario with dense traffic, which provides limited opportunities for safe merging. 

The results show that MARL-MASS strictly respects the safety constraint based on time headway while improving traffic efficiency through higher average speeds, increased merging percentages, and a lower minimum headway than the baseline MARL-HSS.
The custom reward function improves the stability of MARL policies trained with safety shields. 
Although baseline policies such as unsafe MARL and MARL-CS achieve traffic efficiency comparable to that of MARL-MASS, they lack safety guarantees.
In contrast, the proposed lane change controller enables collaboration and effectively balances conflicting objectives to change lanes efficiently in a congested merging scenario while ensuring safety. 

In future work, the MARL-MASS should be extended to other scenarios where congestion is common, such as double merging and traffic intersections. Moreover, we plan to extend MARL-MASS to mixed traffic scenarios.     


\section*{Acknowledgment}
The authors thank the editors and anonymous reviewers for their valuable comments. 
This publication has emanated from research conducted with the financial support of Taighde Éireann – Research Ireland under Grant numbers 18/CRT/6222 at the Centre for Research Training in Advanced Networks for Sustainable Societies (ADVANCE CRT) and 13/RC/2077\_P2 at CONNECT: the Research Ireland Centre for Future Networks. 


\bibliographystyle{IEEEtran}

\begin{thebibliography}{00}
\bibitem{shaygan_traffic_2022}
M.~Shaygan, C.~Meese, W.~Li, X.~G. Zhao, and M.~Nejad, ``Traffic prediction using artificial intelligence: {Review} of recent advances and emerging opportunities,'' \emph{Transportation Research Part C: Emerging Technologies}, Dec. 2022.

\bibitem{schrank_urban_2024}
D.~Schrank, L.~Albert, K.~Jha, and B.~Eisele, ``Urban {Mobility} {Report} 2023,'' Texas A\&M Transportation Institute, Tech. Rep., Jun. 2024. 

\bibitem{garg_can_2023}
M.~Garg and M.~Bouroche, ``Can {Connected} {Autonomous} {Vehicles} {Improve} {Mixed} {Traffic} {Safety} {Without} {Compromising} {Efficiency} in {Realistic} {Scenarios}?'' \emph{IEEE Transactions on Intelligent Transportation Systems}, Jun. 2023.

\bibitem{hegde_design_2022}
B.~Hegde and M.~Bouroche, ``{Design of {AI}-based lane changing modules in connected and autonomous vehicles: a survey},'' in \emph{{Twelfth {International} {Workshop} on {Agents} in {Traffic} and {Transportation}}}, Vienna, 2022.

\bibitem{albrecht_multi-agent_2024}
S.~V. Albrecht, F.~Christianos, and L.~Schäfer, \emph{{Multi-{Agent} {Reinforcement} {Learning}: {Foundations} and {Modern} {Approaches}}}.\hskip 1em plus 0.5em minus 0.4em\relax MIT Press, Dec. 2024.

\bibitem{yu_distributed_2020}
C.~Yu, X.~Wang, X.~Xu, M.~Zhang, H.~Ge, J.~Ren, L.~Sun, B.~Chen, and G.~Tan, ``Distributed {Multiagent} {Coordinated} {Learning} for {Autonomous} {Driving} in {Highways} {Based} on {Dynamic} {Coordination} {Graphs},'' \emph{IEEE Transactions on Intelligent Transportation Systems}, Feb. 2020.

\bibitem{dong_space-weighted_2021}
J.~Dong, S.~Chen, Y.~Li, R.~Du, A.~Steinfeld, and S.~Labi, ``{Space-weighted information fusion using deep reinforcement learning: {The} context of tactical control of lane-changing autonomous vehicles and connectivity range assessment},'' \emph{{Transportation Research Part C: Emerging Technologies}}, Jul. 2021. 

\bibitem{chen_graph_2021}
S.~Chen, J.~Dong, P.~Y.~J. Ha, Y.~Li, and S.~Labi, ``{Graph neural network and reinforcement learning for multi-agent cooperative control of connected autonomous vehicles},'' \emph{{Computer-Aided Civil and Infrastructure Engineering}}, 2021.

\bibitem{zhou_multi-agent_2022}
W.~Zhou, D.~Chen, J.~Yan, Z.~Li, H.~Yin, and W.~Ge, ``{Multi-agent reinforcement learning for cooperative lane changing of connected and autonomous vehicles in mixed traffic},'' \emph{{Autonomous Intelligent Systems}}, Mar. 2022.

\bibitem{hou_decentralized_2021}
Y.~Hou and P.~Graf, ``Decentralized {Cooperative} {Lane} {Changing} at {Freeway} {Weaving} {Areas} {Using} {Multi}-{Agent} {Deep} {Reinforcement} {Learning},'' Oct. 2021, arXiv: 2110.08124.

\bibitem{chen_deep_2023}
D.~Chen, M.~R. Hajidavalloo, Z.~Li, K.~Chen, Y.~Wang, L.~Jiang, and Y.~Wang, ``Deep {Multi}-{Agent} {Reinforcement} {Learning} for {Highway} {On}-{Ramp} {Merging} in {Mixed} {Traffic},'' \emph{IEEE Transactions on Intelligent Transportation Systems}, 2023.

\bibitem{hegde_multi-agent_2024}
B.~Hegde and M.~Bouroche, ``{Multi-agent reinforcement learning for safe lane changes by connected and autonomous vehicles: {A} survey},'' \emph{{AI Communications}}, Jan. 2024.

\bibitem{cheng_end--end_2019}
R.~Cheng, G.~Orosz, R.~M. Murray, and J.~W. Burdick, ``{End-to-{End} {Safe} {Reinforcement} {Learning} through {Barrier} {Functions} for {Safety}-{Critical} {Continuous} {Control} {Tasks}},'' \emph{{Proceedings of the AAAI Conference on Artificial Intelligence}}, Jul. 2019.

\bibitem{zhang_control_2025}
C.~Zhang, L.~Dai, H.~Zhang, and Z.~Wang, ``Control {Barrier} {Function}-{Guided} {Deep} {Reinforcement} {Learning} for {Decision}-{Making} of {Autonomous} {Vehicle} at {On}-{Ramp} {Merging},'' \emph{IEEE Transactions on Intelligent Transportation Systems}, 2025.

\bibitem{hegde_safe_2025}
B.~Hegde and M.~Bouroche, ``Safe and {Efficient} {CAV} {Lane} {Changing} using {Decentralised} {Safety} {Shields},'' in \emph{2025 {IEEE} {Intelligent} {Vehicles} {Symposium} ({IV})}, Jun. 2025.

\bibitem{marinescu_-ramp_2012}
D.~Marinescu, J.~Čurn, M.~Bouroche, and V.~Cahill, ``On-ramp traffic merging using cooperative intelligent vehicles: {A} slot-based approach,'' in \emph{2012 15th {International} {IEEE} {Conference} on {Intelligent} {Transportation} {Systems}}, Sep. 2012.

\bibitem{jing_cooperative_2019}
S.~Jing, F.~Hui, X.~Zhao, J.~Rios-Torres, and A.~J. Khattak, ``Cooperative {Game} {Approach} to {Optimal} {Merging} {Sequence} and on-{Ramp} {Merging} {Control} of {Connected} and {Automated} {Vehicles},'' \emph{IEEE Transactions on Intelligent Transportation Systems}, Nov. 2019.

\bibitem{valiente_robustness_2022}
R.~Valiente, B.~Toghi, R.~Pedarsani, and Y.~P. Fallah, ``Robustness and {Adaptability} of {Reinforcement} {Learning}-{Based} {Cooperative} {Autonomous} {Driving} in {Mixed}-{Autonomy} {Traffic},'' \emph{IEEE Open Journal of Intelligent Transportation Systems}, vol.~3, 2022.

\bibitem{liang_mas-based_2023}
J.~Liang, Y.~Li, G.~Yin, L.~Xu, Y.~Lu, J.~Feng, T.~Shen, and G.~Cai, ``A {MAS}-{Based} {Hierarchical} {Architecture} for the {Cooperation} {Control} of {Connected} and {Automated} {Vehicles},'' \emph{IEEE Transactions on Vehicular Technology}, vol.~72, Feb. 2023.

\bibitem{butler_collaborative_2024}
B.~A. Butler, C.~H. Leung, and P.~E. Paré, ``Collaborative {Safe} {Formation} {Control} for {Coupled} {Multi}-{Agent} {Systems},'' in \emph{2024 {European} {Control} {Conference} ({ECC})}, Jun. 2024.

\bibitem{tan_distributed_2022}
X.~Tan and D.~V. Dimarogonas, ``Distributed {Implementation} of {Control} {Barrier} {Functions} for {Multi}-agent {Systems},'' \emph{IEEE Control Systems Letters}, 2022.

\bibitem{paden_survey_2016}
B.~Paden, M.~Čáp, S.~Z. Yong, D.~Yershov, and E.~Frazzoli, ``A {Survey} of {Motion} {Planning} and {Control} {Techniques} for {Self}-{Driving} {Urban} {Vehicles},'' \emph{IEEE Transactions on Intelligent Vehicles}, Mar. 2016.

\bibitem{polack_kinematic_2017}
P.~Polack, F.~Altché, B.~d'Andréa Novel, and A.~de~La~Fortelle, ``The kinematic bicycle model: {A} consistent model for planning feasible trajectories for autonomous vehicles?'' in \emph{2017 {IEEE} {Intelligent} {Vehicles} {Symposium} ({IV})}, Jun. 2017.

\bibitem{ames_control_2019}
A.~D. Ames, S.~Coogan, M.~Egerstedt, G.~Notomista, K.~Sreenath, and P.~Tabuada, ``Control {Barrier} {Functions}: {Theory} and {Applications},'' in \emph{2019 18th {European} {Control} {Conference} ({ECC})}, Jun. 2019.

\bibitem{han_multi-agent_2023}
S.~Han, S.~Zhou, J.~Wang, L.~Pepin, C.~Ding, J.~Fu, and F.~Miao, ``A {Multi}-{Agent} {Reinforcement} {Learning} {Approach} for {Safe} and {Efficient} {Behavior} {Planning} of {Connected} {Autonomous} {Vehicles},'' \emph{IEEE Transactions on Intelligent Transportation Systems}, 2023.

\bibitem{leurent_environment_2018}
E.~Leurent, ``An {Environment} for {Autonomous} {Driving} {Decision}-{Making},'' 2018, publication Title: GitHub repository. [Online]. Available: \url{https://github.com/eleurent/highway-env}
\end{thebibliography}

\end{document}